\documentclass[11pt,a4paper]{article}
\usepackage[hyperref]{acl2020}
\usepackage{times}
\usepackage{latexsym}

\aclfinalcopy

\usepackage{graphicx}
\usepackage{subfig}
\usepackage{amssymb}
\usepackage{tikz}
\usepackage[ruled,vlined]{algorithm2e}
\usetikzlibrary{bayesnet}

\usepackage{multirow}
\usepackage{booktabs}
\usepackage{wrapfig}

\usepackage{microtype}
\newcommand{\software}[1]{\texttt{#1}}

\title{Unsupervised Ranking and Aggregation of Label Descriptions for Zero-Shot Classifiers}

\author{
    Angelo Basile\\
    {\small \texttt{angelo.basile@symanto.com}}\\
    Symanto Research\\
    Universitat Politècnica de València\\Spain\\
    \And
    Marc Franco-Salvador\\
    {\small \texttt{marc.franco@symanto.com}}\\
    Symanto Research\\Spain\\
    \And
    Paolo Rosso\\
    {\small \texttt{prosso@dsic.upv.es}}\\
    PRHLT Research Center\\Universitat Politècnica de València\\Spain\\
    }

\begin{document}
\maketitle              
\begin{abstract}
Zero-shot text classifiers based on label descriptions embed an input text and a set of labels into the same space: measures such as cosine similarity can then be used to select the most similar label description to the input text as the predicted label. In a true zero-shot setup, designing good label descriptions is challenging because no development set is available.
Inspired by the literature on Learning with Disagreements, we look at how probabilistic models of repeated rating analysis can be used for selecting the best label descriptions in an unsupervised fashion. We evaluate our method on a set of diverse datasets and tasks (sentiment, topic and stance). Furthermore, we show that multiple, noisy label descriptions can be aggregated to boost the performance.

\end{abstract}
\section{Introduction}
\begin{figure*}[!t]
    \centering
    \includegraphics[width=\textwidth]{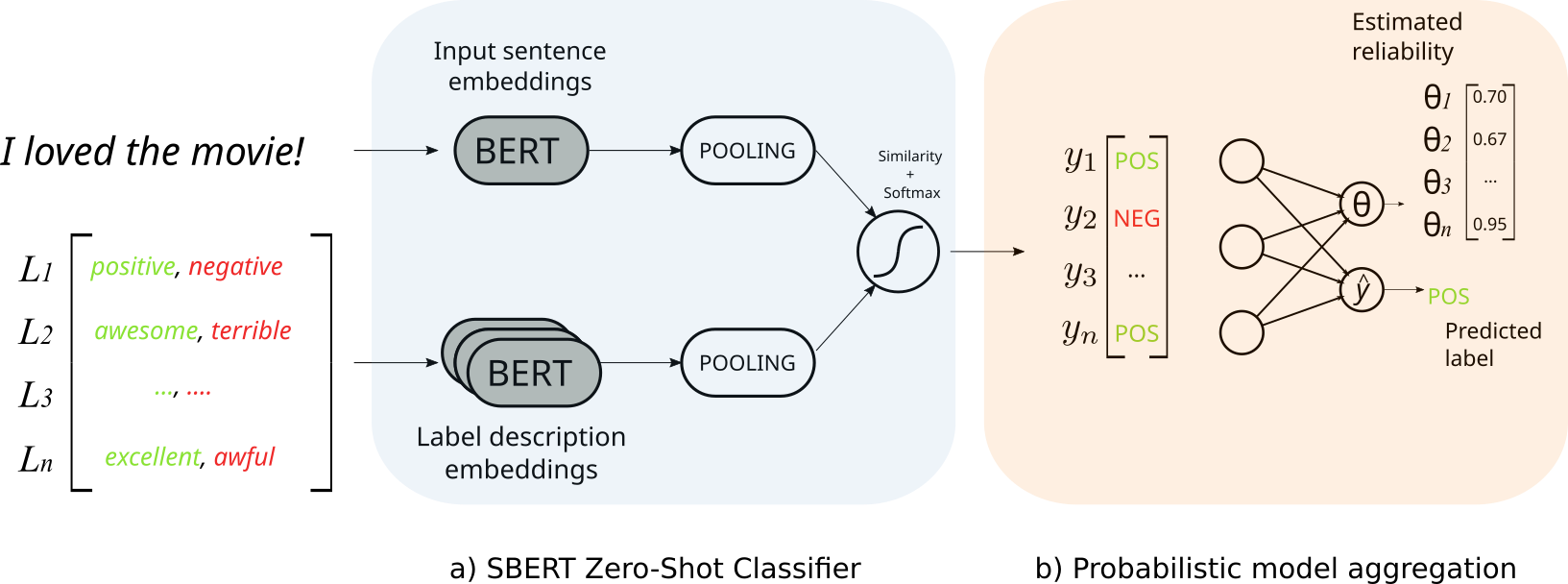}
    \caption{Overview of the proposed method for a sentiment analysis task with two possible output labels, negative (\textsc{NEG}) and positive (\textsc{POS}). An unlabeled corpus with $T$ documents is embedded through \textit{a}) SBERT together with a set of $n$ label descriptions $L$. A softmax classifier on top of the SBERT cosine similarity scores provides $n$ predicted labels $y$, one for each item in the label description set. These predictions are then passed as input to \textit{b}) the probabilistic aggregator, which outputs $\hat{y}$, a best guess of a single final predicted label for each document $T$ and $\theta$, a reliability score for the whole corpus for each label description $y$.}
    \label{fig:pipeline}
\end{figure*}

Recently, large Language Models (LMs) such as BERT \cite{devlin-etal-2019-bert} have pushed the boundaries of NLP systems and have enabled a transition from the supervised learning paradigm, where an input text is processed together with a ground-truth label, to a \textit{pre-train, prompt and predict} paradigm \cite{liu2021pre}, where a pre-trained LM is fed the input data to be processed and a description of the task to be performed. This paradigm shift has lead to zero-shot models that require no ground-truth labels. With a good task description, zero-shot models have been shown to be effective at many challenging NLP tasks \cite{brown2020language}. However, LMs are highly sensitive to how a task description is framed \cite{jiang-etal-2020-know} and, without a large development set, finding a good task description is hard. In this work we address this problem, focusing on zero-shot models based on Siamese BERT-Networks (SBERT) \cite{reimers-gurevych-2019-sentence}. These networks embed both the input text and a description of the target labels in the same semantic space using a pre-trained LM; in this space, similarity measures are then applied to map the most similar label description to the most probable labels. As with prompting, a good label description is key to obtaining good performance. For instance, in the context of binary sentiment analysis, the words \textit{awesome, perfect, great}  and \textit{bad, terrible, awful} are all potentially good descriptions for the labels \textit{positive} and \textit{negative}, respectively. How do we filter out the sub-optimal descriptions without having access to labelled data? 
We study the Learning with Disagreements literature \cite{uma-etal-2021-semeval}, particularly, the item-response class of models, and show that methods developed for analysing crowd-sourced annotations can be transferred to the problem of description selection for zero-shot models. It has been shown that these models achieve high performance in two tasks related to disagreement analysis. First, they usually outperform majority voting at retrieving the gold truth. Second, they can identify which annotators are more reliable and which are spammers. In this work, we look at how we can use these models in the domain of model-based zero-shot classification. Figure \ref{fig:pipeline} shows an illustration of the proposed method.

\section{Zero-Shot Classification with SBERT}

Zero-Shot classification models can tackle a task using no training data. Starting from a pre-trained LM, different architectures enable zero-shot classification using different strategies, such as prompting \cite{brown2020language,liu2021pre} or Natural Language Inference (NLI) for label entailment \cite{yin-etal-2020-universal}. In this work, we focus on zero-shot classifiers based on Siamese Networks. They have recently been shown to perform on par with other methods while being highly efficient at inference time \cite{labeltuning2022, chu-etal-2021-unsupervised}.
As discussed in \cite{reimers-gurevych-2019-sentence}, pre-trained LMs such as BERT can be modified to use Siamese networks and encode two inputs independently: when this architecture is coupled with a symmetric score function, the inference runtime requires only $O(n)$ computations for $n$ instances regardless of the number of labels.
In contrast, the runtime of standard cross-attention encoders would scale linearly with the number of labels. 
While the natural applications of SBERT Networks are clustering and search-related tasks, they can be used for zero-shot classification by providing the text to be classified as a first input and the possible labels (or label descriptions) as the second input: the output is a matrix of similarity measures, which can be transformed in a probability distribution through a softmax function application. For the details on the SBERT classification architecture, we refer to the original work \cite{reimers-gurevych-2019-sentence}.

\paragraph{Label Descriptions}

Label descriptions are the key component that make it possible to turn a semantic similarity model into a zero-shot classifier. We experiment with four sets of label descriptions. First, we define as a baseline a \textit{identity hypothesis} (IH) label, which we set to be equal to the class name it describes: for example, in the context of sentiment analysis, we define the words \textit{positive} and \textit{negative} as label descriptions for the class \textsc{positive} and \textsc{negative}, respectively. 
Second, as a first source of variation, we experiment with a set of \textit{patterns} for turning a identity hypothesis label into a proper sentence: for example, for the IMDB movie review dataset, we experiment with the pattern \textit{The movie is \{}\texttt{positive}, \texttt{negative}\textit{\}}. Third, for each dataset we manually write multiple variations over the identity hypothesis (e.g., \textit{\{positive, negative\}} $\to$ \textit{\{great, terrible\}}). Finally, we experiment with automatically generated variations of the identity hypothesis: under the assumption that the representation of a word in different languages can be a useful source of additional information for a multilingual encoder, we use a machine translation system to automatically translate the identity hypothesis labels into three different languages, which can be directly fed to a multilingual pretrained SBERT model (see Section \ref{sec:experiments} for more details).
Table \ref{tab:label-desc} shows the label descriptions used for the IMDB movie review dataset.

\begin{table}[!t]
    \centering

    \setlength{\tabcolsep}{1em}
    
    \scalebox{0.9}{
    \begin{tabular}{lccc}
        \toprule
        &&\textsc{positive}&\textsc{negative}\\
        \cmidrule{1-4}
         \textsc{IH}&& positive&negative\\
         \midrule
         \multirow{3}{2em}{\textsc{manual}}
         &&great&terrible\\
         &&really great&really terrible\\ 
         &&a masterpiece&awful\\
         \cmidrule{3-4}
         \multirow{3}{1em}{\textsc{AUTO}}
         &&optimo&terribile\\
         &&grande&terrivel\\
         &&genial&negativo\\
         \midrule
         \multirow{5}{1em}{\textsc{pattern}}
         &\multicolumn{3}{c}{\{\}}\\
         &\multicolumn{3}{c}{It was \{\}}\\
         &\multicolumn{3}{c}{All in all, it was \{\}.}\\
         &\multicolumn{3}{c}{Just \{\}!}\\
         &\multicolumn{3}{c}{The movie is \{\}.}\\
         \bottomrule
    \end{tabular}}
        \caption{Overview of the label descriptions used for the IMDB movie review dataset. The identity hypothesis (IH) labels (\textit{positive, negative}) describe the class \texttt{positive} and \texttt{negative} respectively. The rows \textsc{Manual} and \textsc{Pattern} have been manually compiled, while row \textsc{Auto} shows the label descriptions generated automatically by translating the null hypothesis.
    }
    \label{tab:label-desc}
\end{table}

\section{Bayesian Label Description Analysis}
\label{sec:mace}

\begin{figure}[!t]
    \centering
         \begin{tikzpicture}
        
          \node[obs]                (y) {$y_{i,n}$};
          \node[latent, left=of y]  (g) {$\mathbf{G_{i}}$};
          \node[latent, above=of y] (b) {$\mathbf{B_{i,n}}$};
          \node[latent, right=of b] (theta)
          {$\theta$};
          \node[latent, right=of y] (xi)
          {$\xi$};
        
          \edge {b,g} {y} ; %
          \edge {theta} {b} ; %
          \edge {xi} {y} ; %
        
          \plate {annotators} {(b)(y)} {$N$} ;
          \plate {instances} {(annotators)(g)(y)(annotators.north west)(annotators.south west)} {$I$} ;
          \plate {parameters} {(theta)(xi)}{$J$};
        
        \end{tikzpicture}
   \caption{The model plate for MACE. Given $I$ instances and $N$ annotators, the observed label $y_{i,n}$ is dependent on the gold label $G_i$ and $B_{i,n}$, which models the behaviour of annotator $n$ on instance $i$. }
    \label{fig:mace-plate}
\end{figure}
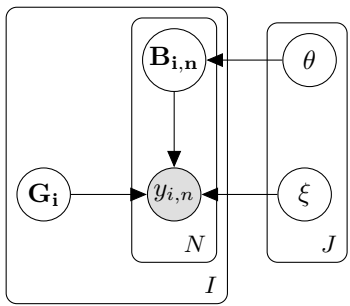
\paragraph{From Crowdsourcing to Zero-Shot Classifiers.}

Bayesian inference provides a natural framework for dealing with multiple sources of uncertain information. This framework is successfully used in the context of analysing crowdsourced annotations as a robust alternative to: \textit{i}) inter-annotator agreement metrics for identifying biases and potential reliability issues in the annotation process, and \textit{ii})  majority voting for retrieving the gold truth label. 
In this work, we argue that this framework can be directly applied to the problem of prompt and label description selection and we use the word \textit{annotator} to denote human annotators, zero-shot classifiers and actual label descriptions.

In NLP, a popular Bayesian annotation model is the Multi-Annotator Competence Estimation (MACE) model \cite{hovy-etal-2013-learning}.
MACE is a generative, unpooled model. It is  \textit{generative} because it can generate a dataset starting from a set of priors, and the assumptions that produce a specific outcome can be represented as a graph, as shown in Figure \ref{fig:mace-plate}. 
Thanks to its \textit{unpooled} structure, it models each annotator (or label description) independently and  provides a trustworthiness score $\theta_j$ for each annotator $j$: we use this parameter for ranking the label descriptions assuming that high $\theta$ values lead to higher f1-scores.
Given $n$ label descriptions and $i$ instances, the zero-shot model outputs $n$ predicted labels $y$: MACE models each label $y_{i,n}$ for instance $i$ from a label description $n$ as being dependent on the true, unobserved gold label $G_i$ and the behaviour $B_{i,n}$ of the zero-shot model with label description $i$ on the instance $n$. The variable $B$ was originally introduced in MACE for modelling the spamming behaviour of crowd-workers.
Key of this work is that, in the context of zero-shot classifiers, $B$ correlates strongly with the true f1-score of different label descriptions.
As a consequence, it can be used to rank the label descriptions and eventually discard the sub-optimal ones. We refer to the original MACE paper \cite{hovy-etal-2013-learning} for additional details on the model.

\section{Experiments and Results}
\label{sec:experiments}

\begin{table}[!t]
    \centering

    {\setlength{\tabcolsep}{8pt}
    
    \begin{tabular}{lcr}
    \toprule
    {} &  $\rho(\theta, f1)$ &  $\rho(\kappa, f1)$ \\
    \midrule
    Ag News          &   \textbf{0.58} &   0.38 \\
    Cola             &   \textbf{0.12} &   0.08 \\
    Imdb             &   \textbf{0.94} &   0.87 \\
    StanceCat        &   \textbf{0.66} &   0.39 \\
    SubJ             &   0.09 &   \textbf{0.20} \\
    Yelp    &   \textbf{0.83} &   0.76 \\
    Yelp Full &   0.16 &   \textbf{0.43} \\
    \bottomrule
    \end{tabular}}
    \caption{Spearman $\rho$ rank correlation with true f1-score for both the $\theta$ parameter of the MACE model and the baseline Cohen's $\kappa$ score. 
    }
    \label{tab:correlations}
\end{table}

\begin{table*}[!t]
\centering

{\setlength{\tabcolsep}{4pt}
\begin{tabular}{llccccccc}
\toprule
             & &  Ag News &  Cola &  Imdb &  StanceCat &  SubJ &  Yelp &  Yelp Full \\
method & aggregation &          &       &       &            &       &                &                   \\
\midrule
\textsc{IH}& - &     10.9 &  38.7 &  66.2 &       32.4 &  52.1 &           66.8 &              31.3 \\

\midrule
\multirow{2}{*}{\textsc{pattern}} & mace &      8.2 &  41.9 &  68.1 &       30.8 &  54.5 &           73.3 &               7.4 \\
             & majority &      8.1 &  40.6 &  67.2 &       31.4 &  51.7 &           72.3 &              \textbf{34.7} \\

\cline{1-9}
\multirow{2}{*}{\textsc{manual}} & mace &      9.7 &  38.7 &  66.3 &       \textbf{33.1} &  \textbf{55.2} &           69.8 &              33.5 \\
             & majority &      9.9 &  38.7 &  66.2 &       28.0 &  50.0 &           68.3 &              31.0 \\
\cline{1-9}
\multirow{2}{*}{\textsc{auto}} & mace &     10.3 &  \textbf{47.3} &  \textbf{68.7} &       31.5 &  43.3 &           \textbf{74.9} &               7.4 \\
             & majority &     \textbf{24.8} &  47.1 &  66.5 &       31.0 &  45.7 &           72.5 &              34.0 \\
\bottomrule
\end{tabular}}

\caption{Macro-averaged F1 scores for the experiments with label aggregations.}
\label{tab:results}
\end{table*}

\subsection{Setup}

For implementing the zero-shot classification module, we use the Python package \software{sentence-transformers} \cite{reimers-gurevych-2019-sentence}. For our experiments on the English corpora, we use the pre-trained model \software{paraphrase-MiniLM-L3-v2} \cite{reimers-gurevych-2021-curse}, which has been trained on a variety of different datasets. We evaluate our proposed method on a battery of popular text classification datasets: IMDB \cite{maas-etal-2011-learning}, Yelp Review \cite{zhang2015character}, Yelp Polarity Review \cite{zhang2015character}, AG's News Corpus \cite{AgNews}, Cola \cite{warstadt2018neural}. In addition, we include StanceCat \cite{taule2017overview}, a stance detection dataset for which non-aggregated annotations are available. For the identity hypothesis (IH) label descriptions, we re-use the prompts that we could find in the literature on prompting \cite{yin-etal-2019-benchmarking,wang2021entailment,labeltuning2022} and manually crafted the rest.\footnote{The complete of the label descriptions used can be found at \url{https://github.com/anbasile/zsla/}}
We manually wrote the variations on the identity hypothesis based on our intuitions.
For the automatic generation of new label descriptions, we translated the English identity hypothesis into French, Italian and Spanish using a pre-trained MarianMT model \cite{junczys-dowmunt-etal-2018-marian} through the \software{transformers} library \cite{wolf-etal-2020-transformers}.
Specifically, we used the \software{opus-mt-en-roa} model \cite{TiedemannThottingal:EAMT2020}. 

We conduct our experiments using the Stan-based implementation \cite{paun-etal-2018-comparing} of MACE, trained with VB.

\subsection{Results}

\paragraph{Ranking.} 

Table \ref{tab:correlations} shows the rank correlation scores with the true f1-score for different datasets. As a baseline, we use the average of Cohen's $\kappa$ \cite{artstein-poesio-2008-survey} computed between each pair of label descriptions.
For most of the datasets, the MACE's $\theta$ parameter, which models the trustworthiness of an annotator, outperforms the baseline.
Medium to strong correlation between MACE's $\theta$ parameters and the true f1-score suggests that a model-based analysis using zero-shot classifiers can be used to effectively select the best performing label descriptions and discard the sub-optimal ones, i.e., by ranking the different label descriptions according to the $\theta$ values, the low-scoring labels can be safely left out. 

\paragraph{Aggregation.} Table \ref{tab:results} shows the results of the label aggregation experiments. In all of the cases,  aggregating multiple label descriptions outperforms the identity hypothesis (\textsc{IH}) baseline. 
In addition, MACE usually outperforms majority voting, excluding the cases where the label space contains more than two labels (i.e., Ag News and Yelp Full).
On average, the automatically generated label descriptions outperform both the manually written label descriptions and the pattern variations: this suggests that human involvement is not necessarily needed for finding better performing label descriptions.

\section{Related Work}

The idea of using the meaning of a category for building dataless classification systems has first been explored already in pre-neural times \cite{chang2008importance}. 
Within the \textit{pretrain, prompt and predict} paradigm \cite{liu2021pre}, automatic prompt generation and ensembling has been investigated in \citet{jiang-etal-2020-know}. \citet{schick-schutze-2021-exploiting} train a classifier on top of the soft-labels provided by an ensemble of zero-shot models for successfully cancelling the effect of poorly-performing prompts.

The idea of modelling ambiguity and disagreement in annotation as signal more than noise, has recently gained traction in the NLP community \cite{uma-etal-2021-semeval,plank-etal-2014-linguistically,fornaciari-etal-2021-beyond}.
The closest source to our paper is probably \cite{simpson2013dynamic}, who use a Bayesian model to combine multiple weak classifiers in a better performing system. \cite{paun-etal-2018-comparing,passonneau-carpenter-2014-benefits} highlight the benefits of Bayesian models for NLP specifically.

\section{Conclusion}

We set out to address two research questions: first, we looked at the problem of unsupervised ranking of different label descriptions by their estimated trustworthiness on different text classification datasets; second, we investigated whether  the output of zero-shot models built with different label descriptions can be aggregated in order to obtain overall higher classification performance. We found that Bayesian models of annotations such as MACE can provide a good solution for both problems. Furthermore, we have found that automatically translated label descriptions outperform manually written ones. We focused on Siamese zero-shot models because their inference runtime is not affected by the number of label descriptions.
When put all together, these findings suggest that zero-shot model performance can be improved by automatically generating more label descriptions and aggregating their output with a probabilistic model. 

\section*{Acknowledgements}

We gracefully thank the support of the Pro$^2$Haters - Proactive Profiling of Hate Speech Spreaders (CDTi IDI-20210776), XAI-DisInfodemics: eXplainable AI for disinformation and conspiracy detection during infodemics (MICIN PLEC2021-007681), DETEMP - Early Detection of Depression Detection in Social Media (IVACE IMINOD/2021/72) and DeepPattern (PROMETEO/2019/121) R\&D grants. Grant PLEC2021-007681 funded by MCIN/AEI/ 10.13039/501100011033 and by European Union NextGenerationEU/PRTR.

\bibliographystyle{acl_natbib}
\bibliography{bib/acl_anthology, bib/bibliography}

\end{document}